\documentclass[final]{cvpr}

\usepackage{times}
\usepackage{epsfig}
\usepackage{graphicx}
\usepackage{amsmath}
\usepackage{tabularx}
\usepackage{amssymb}
\usepackage{booktabs}
\usepackage{color}

\usepackage[pagebackref=true,breaklinks=true,colorlinks,bookmarks=false]{hyperref}

\def\cvprPaperID{5682} 
\def\confYear{CVPR 2021}

\begin{document}

\title{DeepOIS: Gyroscope-Guided Deep Optical Image Stabilizer Compensation}

\author{
    Haipeng Li$^{1}$\ \ \ Shuaicheng Liu$^{2,1}$\ \ \ Jue Wang$^{1}$
	\\
	\\
    $^{1}$Megvii Technology \\
    $^{2}$University of Electronic Science and Technology of China \\
}

\maketitle

\begin{abstract}
Mobile captured images can be aligned using their gyroscope sensors. Optical image stabilizer (OIS) terminates this possibility by adjusting the images during the capturing. In this work, we propose a deep network that compensates the motions caused by the OIS, such that the gyroscopes can be used for image alignment on the OIS cameras$\footnote{Code will be available on \href{https://github.com/lhaippp/DeepOIS}{https://github.com/lhaippp/DeepOIS}.}$. To achieve this, first, we record both videos and gyroscopes with an OIS camera as training data. Then, we convert gyroscope readings into motion fields. Second, we propose a Fundamental Mixtures motion model for rolling shutter cameras, where an array of rotations within a frame are extracted as the ground-truth guidance. Third, we train a convolutional neural network with gyroscope motions as input to compensate for the OIS motion. Once finished, the compensation network can be applied for other scenes, where the image alignment is purely based on gyroscopes with no need for images contents, delivering strong robustness. Experiments show that our results are comparable with that of non-OIS cameras, and outperform image-based alignment results with a relatively large margin.
\end{abstract}

\section{Introduction}

Image alignment is a fundamental research problem that has been studied for decades, which has been applied in various applications~\cite{BrownL03,zaragoza2013projective,guo2016joint,WronskiGEKKLLM19,liu2013bundled}. Commonly adopted registration methods include homography~\cite{gao2011constructing}, mesh-based deformation~\cite{zaragoza2013projective,LiuTYSZ16}, and optical flow~\cite{dosovitskiy2015FlowNet,EpicFlow_2015}. These methods look at the image contents for the registration, which often require rich textures~\cite{lin2017direct,zhang2020content} and similar illumination variations~\cite{sun2018pwc} for good results.


In contrast, gyroscopes can be used to align images, where image contents are no longer required~\cite{karpenko2011digital}. The gyroscope in a mobile phone provides the camera 3D rotations, which can be converted into homographies given camera intrinsic parameters for the image alignment~\cite{karpenko2011digital, hartley2003multiple}. In this way, the rotational motions can be compensated. We refer to this as gyro image alignment. One drawback is that translations cannot be handled by the gyro. Fortunately, rotational motions are prominent compared with translational motions~\cite{shan2007rotational}, especially when filming scenes or objects that are not close to the camera~\cite{liu2017hybrid}. 

Compared with image-based methods, gyro-based methods are attractive. First, it is irrelevant to image contents, which largely improves the robustness. Second, gyros are widely available and can be easily accessed on our daily mobiles. Many methods have built their applications based on the gyros~\cite{huang2018online,guse2012gesture,zaki2020study}. 

\begin{figure}[t]
\begin{center}
  \includegraphics[width=1\linewidth]{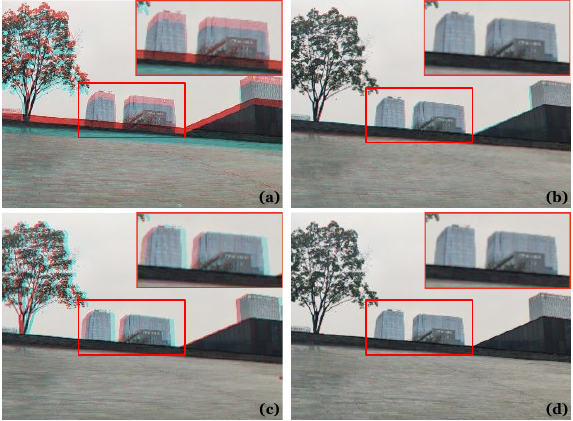}
\end{center}
  \caption{(a) inputs without the alignment, (b) gyroscope alignment on a non-OIS camera, (c) gyroscope alignment on an OIS camera, and (d) our method on an OIS camera. We replace the red channel of one image with that of the other image, where misaligned pixels are visualized as colored ghosts.  The same visualization is applied for the rest of the paper.}
\label{fig:teaser}
\end{figure}

On the other hand, the cameras of smartphones keep evolving, where optical image stabilizer (OIS) becomes more and more popular, which promises less blurry images and smoother videos. It compensates for 2D pan and tilt motions of the imaging device through lens mechanics~\cite{chiu2007optimal,yeom2009optical}. However, OIS terminates the possibility of image registration by gyros. As the homography derived from the gyros is no longer correspond to the captured images, which have been adjusted by OIS with unknown quantities and directions. One may try to read pans and tilts from the camera module. However, this is not easy as it is bounded with the camera sensor, which requires assistance from professionals of the manufacturers~\cite{koo2009optical}.

In this work, we propose a deep learning method that compensates the OIS motion without knowing its readings, such that the gyro can be used for image alignment on OIS equipped cell-phones. Fig.~\ref{fig:teaser} shows an alignment example. Fig.~\ref{fig:teaser} (a) shows two input images. Fig.~\ref{fig:teaser} (b) is the gyro alignment produced by a non-OIS camera. As seen, the images can be well aligned with no OIS interferences. Fig.~\ref{fig:teaser} (c) is the gyro alignment produced by an OIS camera. Misalignments can be observed due to the OIS motion. Fig.~\ref{fig:teaser} (d) represents our OIS compensated result. 

Two frames are denoted as $I_a$ and $I_b$, the motion from gyro between them as $G_{ab}$. The real motion (after OIS adjustment) between two frames is $G'_{ab}$. We want to find a mapping function that transforms $G_{ab}$ to $G'_{ab}$:
\begin{equation}
    \small
    G'_{ab} = f(G_{ab}).
\end{equation}

We propose to train a supervised convolutional neural network for this mapping. To achieve this, we record videos and their gyros as training data. The input motion $G_{ab}$ can be obtained directly given the gyro readings. However, obtaining the ground-truth labels for $G'_{ab}$ is non-trivial. We propose to estimate the real motion from the captured images. If we estimate a homography between them, then the translations are included, which is inappropriate for rotation-only gyros. The ground-truth should merely contain rotations between $I_a$ and $I_b$. Therefore, we estimate a fundamental matrix and decompose it for the rotation matrix~\cite{hartley2003multiple}. However, the cell-phone cameras are rolling shutter (RS) cameras, where different rows of pixels have slightly distinct rotations matrices. In this work, we propose a Fundamental Mixtures model that estimates an array of fundamental matrices for the RS camera, such that rotational motions can be extracted as the ground-truth. In this way, we can learn the mapping function. 

For evaluations, we capture a testing dataset with various scenes, where we manually mark point correspondences for quantitative metrics. According to our experiments, our network can accurately recover the mapping, achieving gyro alignments comparable to non-OIS cameras. In summary, our contributions are:
\begin{itemize}
  \item We propose a new problem that compensates OIS motions for gyro image alignment on cell-phones. To the best of our knowledge, the problem is not explored yet, but important to many image and video applications.   
  \item We propose a solution that learns the mapping function between gyro motions and real motions, where a Fundamental Mixtures model is proposed under the RS setting for the real motions. 
  \item We propose a dataset for the evaluation. Experiments show that our method works well when compared with non-OIS cameras, and outperforming image-based opponents in challenging cases.   
\end{itemize}

\section{Related Work}
\subsection{Image Alignments}
Homography~\cite{gao2011constructing}, mesh-based~\cite{zaragoza2013projective}, and optical flow~\cite{sun2018pwc} methods are the most commonly adopted motion models, which align images in a global, middle, and pixel level. They are often estimated by matching image features or optimize photometric loss~\cite{lin2017direct}. Apart from classical traditional features, such as SIFT~\cite{Lowe04}, SURF~\cite{BayTG06}, and ORB~\cite{RubleeRKB11}, deep features have been proposed for improving robustness, e.g., LIFT~\cite{YiTLF16} and SOSNet~\cite{TianYFWHB19}. Registration can also be realized by deep learning directly, such as deep homography estimation~\cite{le2020deep,zhang2020content}. In general, without extra sensors, these methods align images based on the image contents.  

\subsection{Gyroscopes}
Gyroscope is important in helping estimate camera rotations during mobile capturing. The fusion of gyroscope and visual measurements have been widely applied in various applications, including but not limited to, image alignment and video stabilization~\cite{karpenko2011digital}, image deblurring~\cite{mustaniemi2019gyroscope}, simultaneous localization and mapping (SLAM)~\cite{huang2018online}, gesture-based user authentication on mobile devices~\cite{guse2012gesture}, and human gait recognition~\cite{zaki2020study}. In mobiles, one important issue is the synchronization between the timestamps of gyros and video frames, which requires gyro calibration~\cite{jia2013online}. In this work, we access the gyro data at the Hardware Abstraction Layer (HAL) of the android layout~\cite{siddha2012hardware}, to achieve accurate synchronizations.   

\subsection{Optical Image Stabilizer}
Optical Image Stabilizer (OIS) has been around commercially since the mid-90s~\cite{sato1993control} and becomes more and more popular in our daily cell-phones. Both the image capturing and video recording can benefit from OIS, producing results with less blur and improved stability~\cite{koo2009optical}. It works by controlling the path of the image through the lens and onto the image sensor, which is achieved by measuring the camera shakes using sensors such as gyroscope, and move the lens horizontally or vertically to counteract shakes by electromagnet motors~\cite{chiu2007optimal,yeom2009optical}. Once a mobile is equipped with OIS, it cannot be turn-off easily~\cite{nasiri2012optical}. On one hand, OIS is good for daily users. On the other hand, it is not friendly to mobile developers who need gyros to align images. In this work, we enable the gyro image alignment on OIS cameras.   

\section{Algorithm}
\begin{figure*}[t]
\begin{center}
\includegraphics[width=1\textwidth]{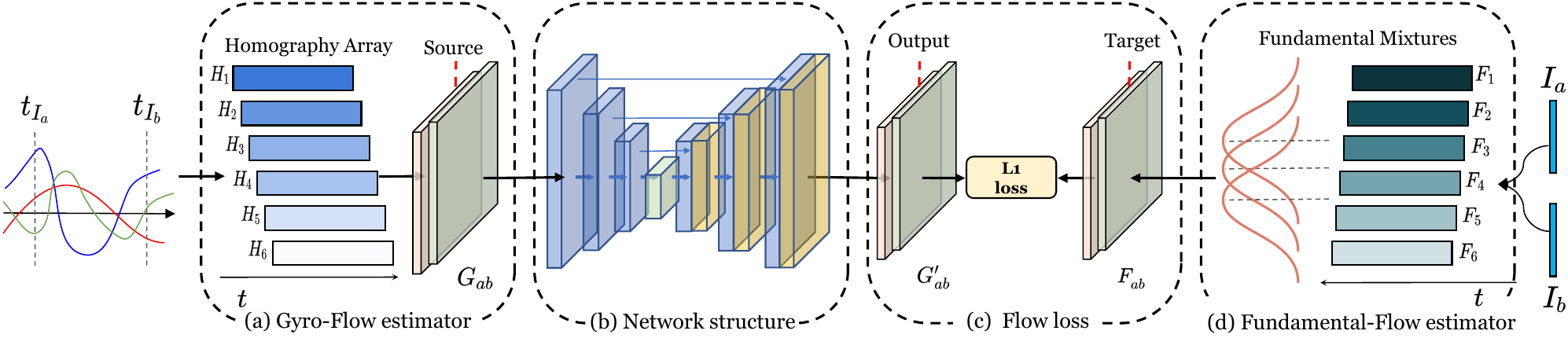}
\end{center}
\caption{The overview of our algorithm which includes (a) gyro-based flow estimator, (d) the fundamental-based flow estimator, and (b) neural network predicting an output flow. For each pair of frames $I_a$ and $I_b$, the homography array is computed using the gyroscope readings from $t_{I_{a}}$ to $t_{I_{b}}$, which is converted into the source motion $G_{ab}$ as the network input. On the other side, we estimate a Fundamental Mixtures model to produce the target flow $F_{ab}$ as the guidance. The network is then trained to produce the output $G'_{ab}$.}
\label{fig:pipeline}
\end{figure*}
Our method is built upon convolutional neural networks. It takes one gyro-based flow $G_{ab}$ from the source frame $I_a$ to the target frame $I_b$ as input, and produces OIS compensated flow $G'_{ab}$ as output. Our pipeline consists of three modules: a gyro-based flow estimator, a Fundamental Mixtures flow estimator, and a fully convolutional network that compensates the OIS motion. Fig.~\ref{fig:pipeline} illustrates the pipeline. First, the gyro-based flows are generated according to the gyro readings (Fig.~\ref{fig:pipeline}(a) and Sec.~\ref{sec:gyroflow}), then they are fed into a network to produce OIS compensated flows $G'_{ab}$ as output (Fig.~\ref{fig:pipeline}(b) and Sec.~\ref{sec:network}). To obtain the ground-truth rotations, we propose a Fundamental Mixtures model, so as to produce the Fundamental Mixtures flows $F_{ab}$ (Fig.~\ref{fig:pipeline} (d) and Sec.~\ref{sec:fundamental}) as the guidance to the network (Fig.~\ref{fig:pipeline} (c)). During the inference, the Fundamental Mixtures model is not required. The gyro readings are converted into gyro-based flows and fed to the network for compensation.

\begin{figure}[t]
\begin{center}
  \includegraphics[width=0.9\linewidth]{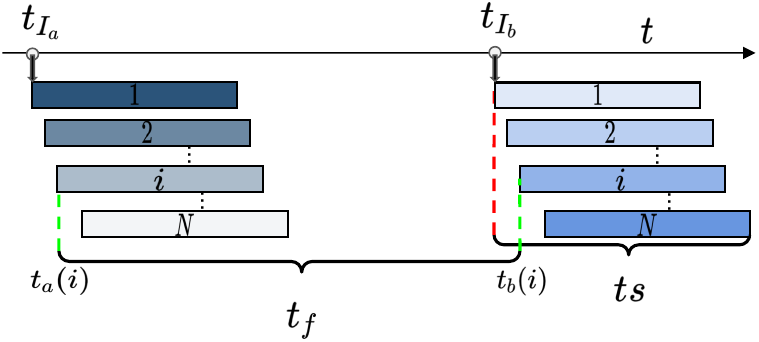}
\end{center}
  \caption{Illustration of rolling shutter frames. $t_{I_a}$ and $t_{I_b}$ are the frame starting time. $t_s$ is the camera readout time and $t_f$ denotes the frame period ($t_f > t_s$). $t_a(i)$ and $t_b(i)$ represent the starting time of patch $i$ in $I_a$ and $I_b$. }
\label{fig:gyro_ts}
\end{figure}

\subsection{Gyro-Based Flow}\label{sec:gyroflow}
We compute rotations by compounding gyro readings consisting of angular velocities and timestamps. In particular, we read them from the HAL of android architecture for synchronization. The rotation vector $n=\left(\omega_{x}, \omega_{y}, \omega_{z}\right) \in \mathbb{R}^{3}$ is computed from gyro readings between frames $I_a$ and $I_b$~\cite{karpenko2011digital}. The rotation matrix $R(t)\in SO(3)$ can be produced according to the Rodrigues Formula~\cite{dai2015euler}. 

If the camera is global shutter, the homography is modeled as:
\begin{equation}
\small
\mathbf{H}(t)=\mathbf{K} \mathbf{R}(t) \mathbf{K}^{-1}\label{eq:globalH},
\end{equation}
\noindent where $K$ is the intrinsic camera matrix and $R(t)$ denotes the camera rotation from $I_a$ to $I_b$.

In an RS camera, every row of the image is exposed at a slightly different time. Therefore, Eq.(\ref{eq:globalH}) is not applicable since every row of the image has slightly different rotation matrices. In practice, assigning each row of pixels with a rotation matrix is unnecessary. We group several consecutive rows into a row patch and assign each patch with a rotation matrix. Fig.~\ref{fig:gyro_ts} shows an example. Let $t_s$ denotes the camera readout time that is the time duration between the exposure of the first row and the last row of pixels. 

\begin{equation}
\small
t_{a}(i)=t_{I}+t_{s} \frac{i}{N},
\end{equation}

\noindent where $t_{a}(i)$ denotes the start of the exposure of the $i$-$th$ patch in $I_a$ as shown in Fig.~\ref{fig:gyro_ts}, $t_{I}$ denotes the starting timestamp of the corresponding frame, $N$ denotes the number of patches per frame. The end of the exposure is:

\begin{equation}
\small
t_{b}(i)=t_{a}(i)+t_{f},
\end{equation}

\noindent where $t_{f}=1/FPS$ is the frame period. Here, the homography between the $i$-th row at frame $I_a$ and $I_b$ can be modeled as:

\begin{equation}
\small
\mathbf{H}=\mathbf{K} \mathbf{R}\left(t_{b}\right) \mathbf{R}^{\top}\left(t_{a}\right) \mathbf{K}^{-1},
\end{equation}
where $\small\mathbf{R}\left(t_{b}\right) \mathbf{R}^{\top}\left(t_{a}\right)$ can be computed by accumulating rotation matrices from $t_a$ to $t_b$.

In our implementation, we divide the image into $6$ patches which computes a homography array containing $6$ horizontal homographies between two consecutive frames. We convert the homography array into a flow field~\cite{mustaniemi2019gyroscope} so that it can be fed as input to a convolutional neural network. For every pixel $p$ in the $I_a$, we have:

\begin{equation}
\small
\mathbf{p}^{\prime}=\mathbf{H}(t) \mathbf{p}, \quad \mathbf{(u,v)}=\mathbf{p}^{\prime}- \mathbf{p},
\label{eq:homo2flow}
\end{equation}
\noindent computing the offset for every pixel produces a gyro-based flow $G_{ab}$.

\subsection{Fundamental Mixtures}\label{sec:fundamental}


Before introducing our model of Fundamental Mixtures, we briefly review the process of estimating the fundamental matrix. If the camera is global-shutter, every row of the frame is imaged simultaneously at a time. Let $p_1$ and $p_2$ be the projections of the 3D point $X$ in the first and second frame, $p_1=P_{1} X$ and $p_2=P_{2} X$, where $P_{1}$ and $P_{2}$ represent the projection matrices. The fundamental matrix satisfies the equation~\cite{hartley2003multiple}:
\begin{equation}
\small
\mathbf{p}_{1}^{T} \mathbf{F} \mathbf{p}_{2}=0,
\label{F-matrix}
\end{equation}
where $p_1=(x_1, y_1, 1)^{T}$ and $p_2=\left(x_1^{\prime}, y_1^{\prime}, 1\right)^{T}$. Let $\mathbf{f}$ be the 9-element vector made up of $F$, then Eq.(\ref{F-matrix}) can be written as:
\begin{equation}
    \small
    \left(x_1^{\prime} x_1, x_1^{\prime} y_1, x_1^{\prime}, y_1^{\prime} x_1, y_1^{\prime} y_1, y_1^{\prime}, x_1, y_1, 1\right) \mathbf{f}=0,
    \label{dlt_F}
\end{equation}
given $n$ correspondences, yields a set of linear equations:
\begin{equation}
\small
\begin{aligned}
&A \mathbf{f}= \left[\begin{array}{ccc}
x_{1}^{\prime}p_1^{T} & y_{1}^{\prime}p_1^{T} & p_1^{T}  \\
\vdots & \vdots & \vdots \\
x_{n}^{\prime}p_n^{T} & y_{n}^{\prime}p_n^{T} & p_n^{T} 
\end{array}\right] \mathbf{f} = 0.
\end{aligned}
\end{equation}
Using at least 8 matching points yields a homogenous linear system, which can be solved under the constraint $\|\mathbf{f}\|_{2}=1$ using the Singular Value Decomposition(SVD) of $A=U D V^{\top}$ where the last column of $V$ is the solution~\cite{hartley2003multiple}.

In the case of RS camera, projection matrices $P_{1}$ and $P_{2}$ vary across rows instead of being frame-global.  Eq.(\ref{F-matrix}) does not hold. Therefore, we introduce Fundamental Mixtures assigning each row patch with a fundamental matrix. 

We detect FAST features~\cite{trajkovic1998fast} and track them by KLT~\cite{shi1994good} between frames. We modify the detection threshold for uniform feature distributions~\cite{grundmann2012calibration,guo2016joint}.

To model RS effects, we divide a frame into $N$ patches, resulting in $N$ unknown fundamental matrices $F_{i}$ to be estimated per frame. If we estimate each fundamental matrix independently, the discontinuity is unavoidable. We propose to smooth neighboring matrices during the estimation as shown in Fig.~\ref{fig:pipeline} (d), where a point $p_1$ not only contributes to its own patch but also influences its nearby patches weighted by the distance. The fundamental matrix for point $p$ is the mixture:

\begin{equation}
\small
F({p_1}) =\sum_{i=1}^{N} F_{i} w_{i}(p_1),
\end{equation}

\noindent where $w_{i}(p_1)$ is the gaussian weight with the mean equals to the middle of each patch, and the sigma $\sigma = 0.001 * h$, where $h$ represents the frame height.

To fit a Fundamental Mixtures $F_{i}$ given a pair of matching points $\left(p_{1}, p_{2}\right)$, we rewrite Eq.(\ref{F-matrix}) as:
\begin{equation}
    \small
    0=\mathbf{p}_{1}^{T} \mathbf{F_{p_1}} \mathbf{p}_{2}=\sum_{i=1}^{N} w_{i}(p_1) \cdot \mathbf{p}_{1}^{T} \mathbf{F_i} \mathbf{p}_{2},
    \label{p1Fp2}
\end{equation}
where $\mathbf{p}_{1}^{T} \mathbf{F}_{\mathbf{k}} \mathbf{p}_{2}$ can be transformed into:
\begin{equation}
\small
A_{p_1}^{i} f_{i} = \left(\begin{array}{ccc}
x_{1}^{\prime}p_1^{T} & y_{1}^{\prime}p_1^{T} & p_1^{T} 
\end{array}\right) f_{i},
\label{dlt_ak}
\end{equation}
where $f_{i}$ denotes the vector formed by concatenating the columns of $F_{i}$. Combining Eq.(\ref{p1Fp2}) and Eq.(\ref{dlt_ak}) yields a $1 \times 9 i$ linear constraint:
\begin{equation}
    \small
    \underbrace{\left(w_{1}(p_1) A_{p_1}^{1} \ldots w_{i}(p_1) A_{p_1}^{i}\right)}_{A_{p_1}} \underbrace{\left(\begin{array}{c}
f_{1} \\
\vdots \\
f_{i}
\end{array}\right)}_{f}=A_{p_1} f=0.
\label{svd_f_mix}
\end{equation}
Aggregating all linear constraints $A_{p_j}$ for every match point $(p_j, p_{j+1})$ yields a homogenous linear system $A\mathbf{f}=0$ that can be solved under the constraint $\|\mathbf{f}\|_{2}=1$ via SVD. 

For robustness, if the number of feature points in one patch is inferior to $8$, Eq.(\ref{svd_f_mix}) is under constrained. Therefore, we add a regularizer to constrain $\lambda\left\|A_{p}^{i}-A_{p}^{i-1}\right\|_{2}=0\small$ to the homogenous system with $\lambda = 1\small$. 


\subsubsection{Rotation-Only Homography}
Given the fundamental matrix $F_i$ and the camera intrinsic $K$, we can compute the essential matrix $E_i$ of the $i$-th patch: $\mathbf{E_i}=\mathbf{K}^{T} \mathbf{F_i} \mathbf{K}\small.$ The essential matrix $E_i$~\cite{hartley2003multiple} can be decomposed into camera rotations and translations, where only rotations $R_i$ are retained. We use $R_i$ to form a rotation-only homography similar to Eq.(\ref{eq:globalH}) and convert the homography array into a flow field as Eq.(\ref{eq:homo2flow}). We call this flow field as Fundamental Mixtures Flow $F_{ab}$. Note that, $R_i$ is spatially smooth, as $F_i$ is smooth, so does $F_{ab}$.

\subsection{Network Structure}\label{sec:network}
The architecture of the network is shown in Fig.~\ref{fig:pipeline} that utilizes a backbone of UNet~\cite{ronneberger2015u} consists of a series of convolutional and downsampling layers with skip connections. The input to the network is gyro-based flow $G'_{ab}$ and the ground-truth target is Fundamental Mixtures flow $F_{ab}$. Our network aims to produce an optical flow of size $H \times W \times 2$ which compensates the motion generated by OIS between $G'_{ab}$ and $F_{ab}$. Besides, the network is fully convolutional which accepts input of arbitrary sizes. 

Our network is trained on $9k$ rich-texture frames with resolution of $360$x$270$ pixels over $1k$ iterations by an Adam optimizer~\cite{kingma2014adam} whose $l_{r}=1.0 \times 10^{-4}$, $\beta_1=0.9$, $\beta_2=0.999$. The batch size is $8$, and for every $50$ epochs, the learning rate is reduced by $20\%$. The entire training process costs about $50$ hours. The implementation is in PyTorch and the network is trained on one NVIDIA RTX 2080 Ti. 

\section{Experimental Results}


\subsection{Dataset}
Previously, there are some dedicated datasets which are designed to evaluate the homography estimation~\cite{zhang2020content} or the image deblurring with the artificial-generated gyroscope-frame pair~\cite{mustaniemi2019gyroscope}, whether none of them combine real gyroscope readings with corresponding video frames. So we propose a new dataset and benchmark GF4. 

\begin{figure}[t]
\begin{center}
   \includegraphics[width=1\linewidth]{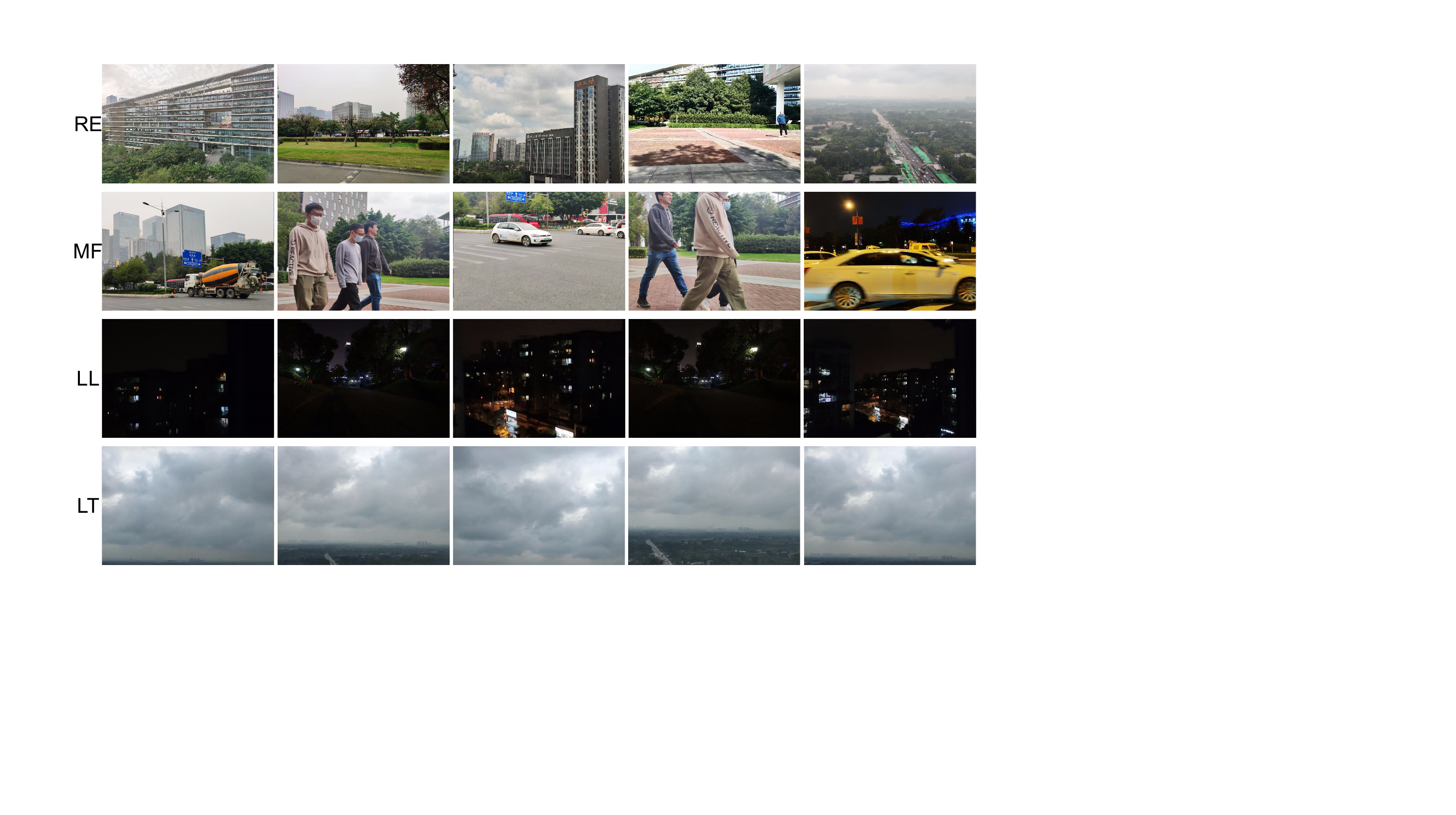}
\end{center}
   \caption{A glance at our evaluation dataset. Our dataset contains 4 categories, regular(RE), low-texture(LT), low-light(LL) and moving-foreground(MF). Each category contains $350$ pairs, a total of $1400$ pairs, with synchronized gyroscope readings.}
\label{fig:dataset}
\end{figure}

\noindent\textbf{Training Set} To train our network, we record a set of videos with the gyroscope readings using a hand-held cellphone. We choose scenes with rich textures so that sufficient feature points can be detected to calculate the Fundamental Mixtures model. The videos last $300$ seconds, yielding $9$,$000$ frames in total. Note that, the scene type is not important as long as it can provide enough features as needed for Fundamental Mixtures estimation. 

\begin{figure}[t]
\begin{center}
   \includegraphics[width=1\linewidth]{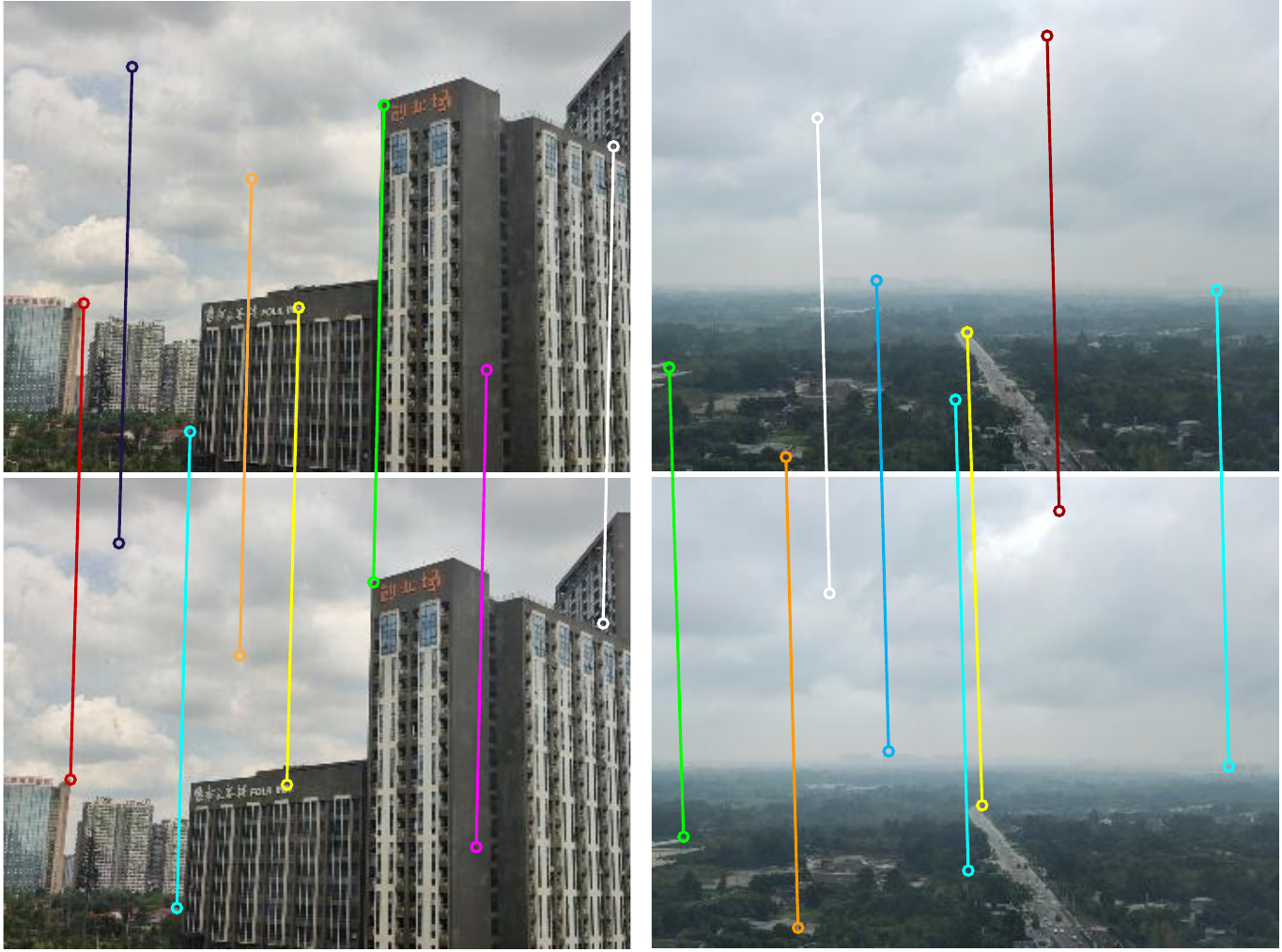}
\end{center}
   \caption{We mark the correspondences manually in our evaluation set for quantitative metrics. For each pair, we mark $6 \sim 8$ point matches. }
\vspace{-5pt}
\label{fig:annotation}
\end{figure}

\noindent\textbf{Evaluation Set} For the evaluation, we capture scenes with different types, to compare with image-based registration methods. Our dataset contains $4$ categories, including regular (RE), low-texture (LT), low-light (LL), and moving-foregrounds (MF) frame-gyroscope pairs. Each scene contains $350$ pairs. So, there are $1400$ pairs in the dataset. We show some examples in Fig.~\ref{fig:dataset}. For quantitative evaluation, we manually mark $6\sim8$ point correspondences per pair, distributing uniformly on frames. Fig.~\ref{fig:annotation} shows some examples. 

\subsection{Comparisons with non-OIS camera}

\begin{figure*}
\begin{center}
\includegraphics[width=1\textwidth]{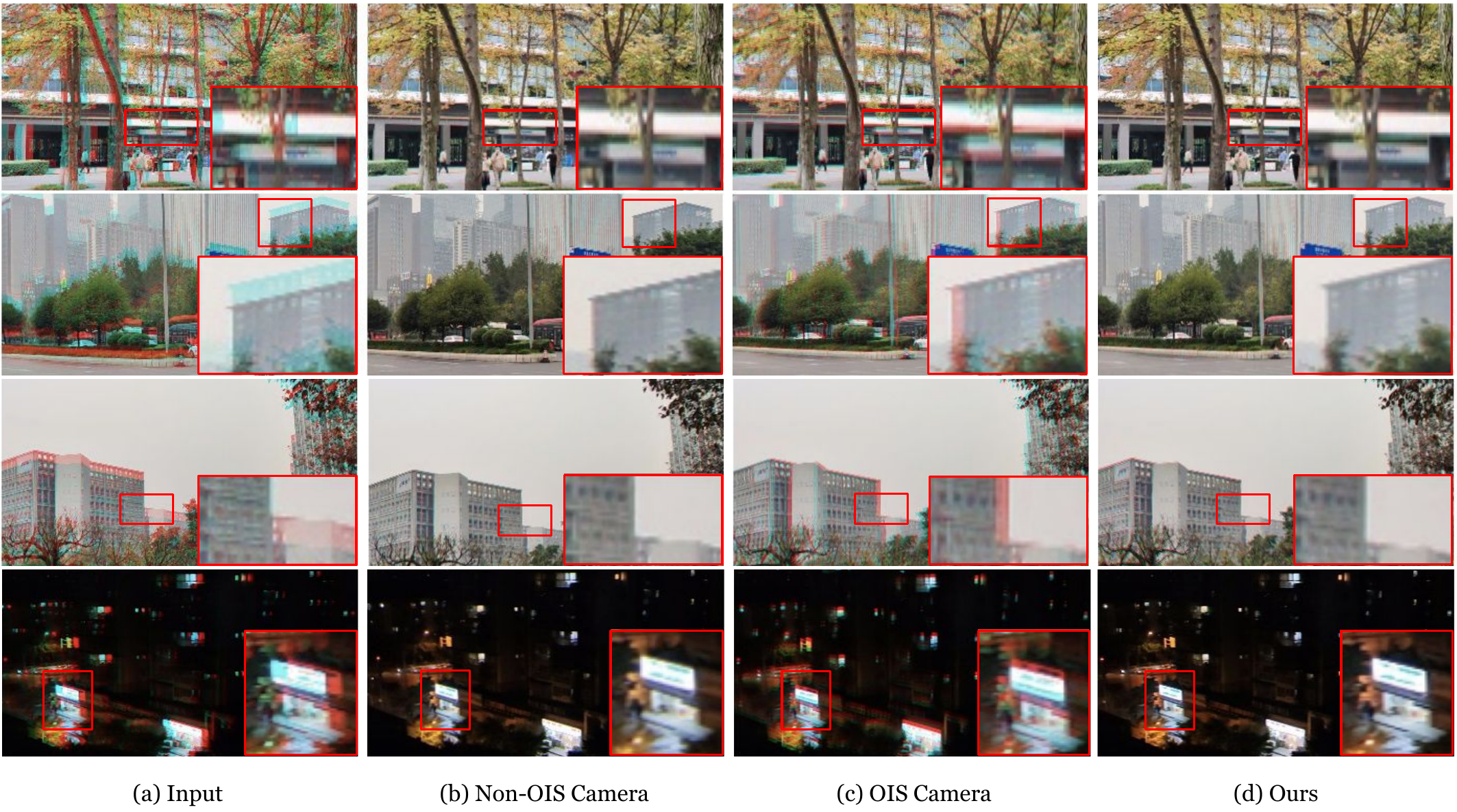}
\end{center}
\vspace{-10pt}
\caption{Comparisons with non-OIS cameras. (a) input two frames. (b) gyro alignment results on the non-OIS camera. (c) gyro alignment results on the OIS camera. (d) our OIS compensation results. Without OIS compensation, clear misalignment can be observed in (c) whereas our method can solve this problem and be comparable with non-OIS results in (b).}
\label{fig:qualitycompare_ois}
\end{figure*}

Our purpose is to enable gyro image alignment on OIS cameras. Therefore, we compare our method with non-OIS cameras. In general, our method should perform equally well as non-OIS cameras, if the OIS motion could be compensated successfully. For comparison, ideally, we should use one camera with OIS turn on and off. However, the OIS cannot be turned off easily. Therefore, we use two cell-phones with similar camera intrinsics, one with OIS and one without, and capture the same scene twice with similar motions. Fig.~\ref{fig:qualitycompare_ois} shows some examples. Fig.~\ref{fig:qualitycompare_ois} (a) shows the input frames. Fig.~\ref{fig:qualitycompare_ois} (b) shows the gyro alignment on a non-OIS camera. As seen, images can be well aligned. Fig.~\ref{fig:qualitycompare_ois} (c) shows the gyro alignment on an OIS camera. Due to the OIS interferences, images cannot be aligned directly using the gyro. Fig.~\ref{fig:qualitycompare_ois} (d) shows our results. With OIS compensation, images can be well aligned on OIS cameras.  

\begin{table}[t]
    \small
    \centering
    \resizebox*{0.45 \textwidth}{!}
    {
    	\begin{tabular}{lccc}
    		\toprule  
    		 &Non-OIS Camera & OIS Camera & Ours\\
    		\midrule  
            Geometry Distance &0.688 & 1.038 & 0.709 \\
    		\bottomrule  
    	\end{tabular}
	}
	\caption{Comparisons with non-OIS camera. }
	\vspace{-10pt}
	\label{tab:non-ois}
\end{table}

We also calculate quantitative values. Similarly, we mark the ground-truth for evaluation. The average geometry distance between the warped points and the manually labeled GT points are computed as the error metric (the lower the better). Table~\ref{tab:non-ois} shows the results. Our result $0.709$ is comparable with non-OIS camera $0.688$ (slightly worse), while no compensation yields $1.038$, which is much higher.  

\subsection{Comparisons with Image-based Methods}

\begin{figure*}[h]
\begin{center}
  \includegraphics[width=1\linewidth]{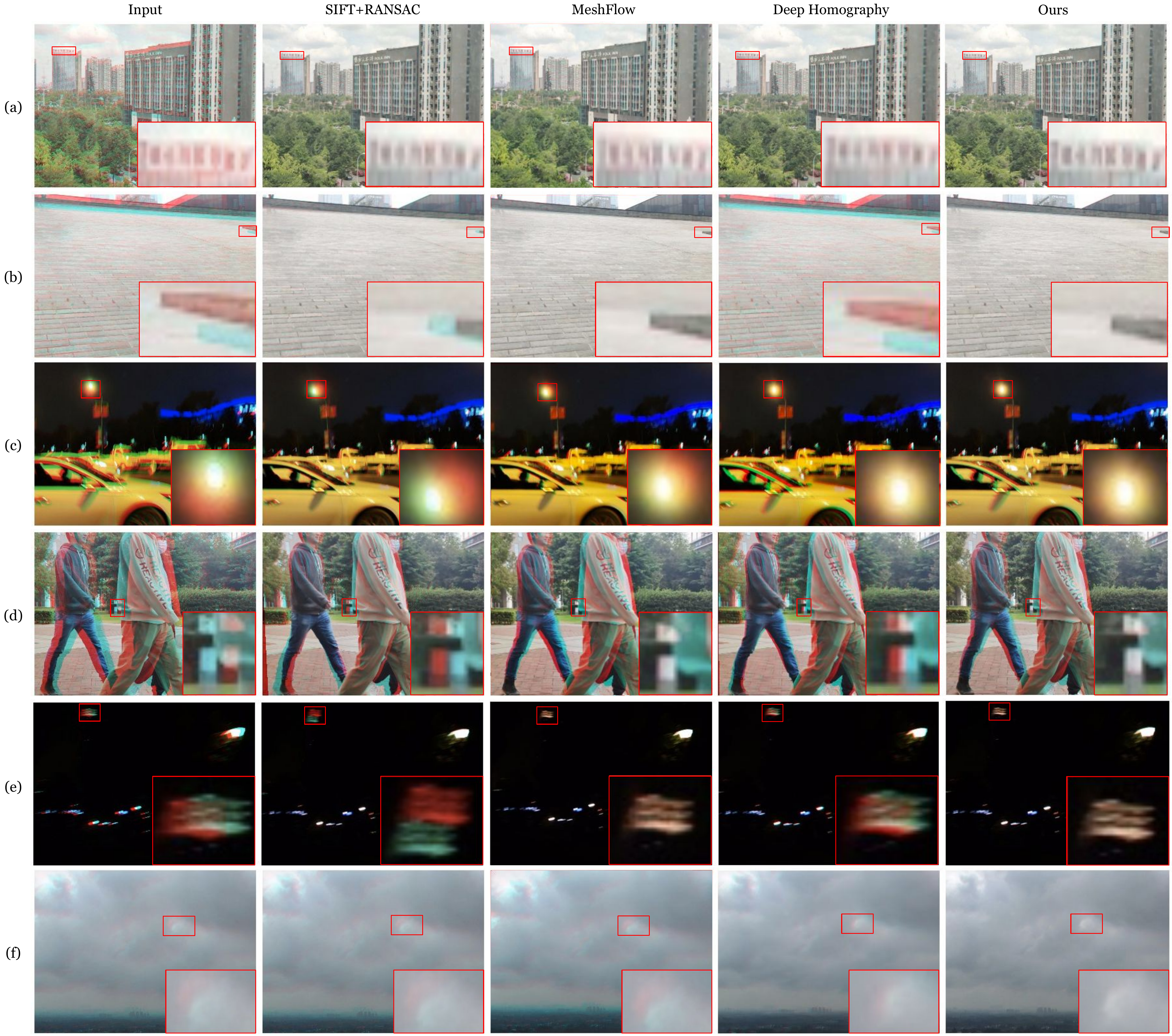}
\end{center}
  \caption{Comparisons with image-based methods. We compare with SIFT~\cite{Lowe04} + RANSAC~\cite{FischlerB81}, Meshflow~\cite{LiuTYSZ16}, and the recent deep homography~\cite{zhang2020content} method. We show examples covering all scenes in our evaluation dataset. Our method can align images robustly while image-based methods contain some misaligned regions.  }
\label{fig:qualitycompare}
\end{figure*}

Although it is a bit unfair to compare with image-based methods as we adopt additional hardware. We desire to show the importance and robustness of the gyro-based alignment, to highlight the importance of enabling this capability on OIS cameras. 

\begin{table*}[h]
\small
\label{tab:results}
	\centering
	\resizebox*{1.0 \textwidth}{!}
	{
	\begin{tabular}{>{\arraybackslash}p{5cm}
    >{\centering\arraybackslash}p{2.4cm}
    >{\centering\arraybackslash}p{2.4cm}
    >{\centering\arraybackslash}p{2.4cm}
    >{\centering\arraybackslash}p{2.4cm}
    >{\centering\arraybackslash}p{2.4cm}}
		\toprule  
		1) &RE& LT& LL& MF& Avg \\ 
		\midrule  
		2) $\mathcal{I}_{3 \times 3}$ & 7.098(+2785.37\%)  & 7.055(+350.80\%)   & 7.035(+519.55\%)   & 7.032(+767.08\%)   & 7.055(+313.97\%)\\
		\midrule  
		3) SIFT~\cite{Lowe04}+ RANSAC~\cite{FischlerB81} & 0.340(+38.21\%)      & 6.242(+298.85\%)   & 2.312(+103.58\%)   & 1.229(+51.54\%)    & 2.531(+48.49\%)  \\
		
		4) SIFT~\cite{Lowe04} + MAGSAC~\cite{barath2019magsac} &\textcolor[rgb]{1,0,0}{0.213(\textminus13.41\%)}      & 5.707(+264.66\%)   & 2.818(+148.17\%)   & \textcolor[rgb]{0,0,1}{0.811(+0.00\%)}     & 2.387(+40.08\%) \\
		
		5) ORB~\cite{rublee2011orb} + RANSAC~\cite{FischlerB81} & 0.653(+165.45\%)     & 6.874(+339.23\%)   & \textcolor[rgb]{0,0,1}{1.136(+0.00\%)}     & 2.27(+179.28\%)   & 2.732(+60.30\%)  \\
		
		6) ORB~\cite{rublee2011orb} + MAGSAC~\cite{barath2019magsac} &0.919(+273.58\%)     & 6.859(+338.27\%)   & 1.335(+17.60\%)    & 2.464(+203.82\%)   & 2.894(+69.83\%) \\
		
		7) SOSNET~\cite{tian2019sosnet} + RANSAC~\cite{FischlerB81} & \textcolor[rgb]{0,0,1}{0.246(+0.00\%)}       & 5.946(+279.94\%)   & 1.977(+74.11\%)    & 0.907(+11.84\%)    & 2.269(+33.14\%)  \\
		
		8) SOSNET~\cite{tian2019sosnet} + MAGSAC~\cite{barath2019magsac} & 0.309(+25.61\%)      & 5.585(+256.87\%)   & 1.972(+73.67\%)    & 1.142(+40.81\%)    & 2.252(+32.14\%) \\
		
		9) SURF~\cite{BayTG06} + RANSAC~\cite{FischlerB81} & 0.343(+39.43\%)      & 3.161(+101.98\%)   & 2.213(+94.89\%)    & 1.420(+75.09\%)    & 1.784(+4.69\%)\\
		
		10) SURF~\cite{BayTG06} + MAGSAC~\cite{barath2019magsac} & 0.307(+24.80\%)      & 3.634(+132.20\%)   & 2.246(+97.78\%)    & 1.267(+56.23\%)    & 1.863(+9.34\%)\\
		
		\midrule  
		11) MeshFlow~\cite{LiuTYSZ16} & 0.843(+242.68\%)     & 7.042(+349.97\%)   & 1.729(+52.27\%)    & 1.109(+36.74\%)    & 2.681(+57.30\%)\\
		
		\midrule  
		12) Deep Homography~\cite{zhang2020content} & 1.342(+445.53\%) & \textcolor[rgb]{0,0,1}{1.565(+0.00\%)} & 2.253(+98.41\%) & 1.657(+104.32\%) & \textcolor[rgb]{0,0,1}{1.704(+0.00\%)}  \\
		
		\midrule  
		13) Ours & 0.609(+147.56\%)     &\textcolor[rgb]{1,0,0}{1.01(\textminus35.27\%)}    &\textcolor[rgb]{1,0,0}{0.637(\textminus43.90\%)}    &\textcolor[rgb]{1,0,0}{0.736(\textminus9.25\%)}     &\textcolor[rgb]{1,0,0}{0.749(\textminus56.07\%)}\\
		\bottomrule  
	\end{tabular}
	}
	\caption{Quantitative comparisons on the evaluation dataset. The best performance is marked in red and the second-best is in blue. }
\label{table:quality}
\end{table*}

\subsubsection{Qualitative Comparisons}
Firstly, we compare our method with one frequently used traditional feature-based algorithm, i.e. SIFT~\cite{Lowe04} and RANSAC~\cite{FischlerB81} that compute a global homography, and another feature-based algorithm, i.e. Meshflow~\cite{LiuTYSZ16} that deforms a mesh for the non-linear motion representation. Moreover, we compare our method with the recent deep homography method~\cite{zhang2020content}.

Fig.~\ref{fig:qualitycompare} (a) shows a regular example where all the methods work well. Fig.~\ref{fig:qualitycompare} (b), SIFT+RANSAC fails to find a good solution, so does deep homography, while Meshflow works well. One possible reason is that a single homography cannot cover the large depth variations.  Fig.~\ref{fig:qualitycompare} (c) illustrates a moving-foreground example that SIFT+RANSAC and Meshflow cannot work well, as few features are detected on the background, whereas Deep Homography and our method can align the background successfully. A similar example is shown in Fig.~\ref{fig:qualitycompare} (d), SIFT+RANSAC and Deep Homography fail. Meshflow works on this example as sufficient features are detected in the background. In contrast, our method can still align the background without any difficulty. Because we do not need the image contents for the registration. Fig.~\ref{fig:qualitycompare}(e) is an example of low-light scenes, and Fig.~\ref{fig:qualitycompare}(f) is a low-texture scene. All the image-based methods fail as no high-quality features can be extracted, whereas our method is robust.

\subsubsection{Quantitative Comparisons}

We also compare our method with other feature-based methods quantitatively, i.e., the geometry distance. For the feature descriptors, we choose SIFT~\cite{Lowe04}, ORB~\cite{rublee2011orb}, SOSNet~\cite{TianYFWHB19}, SURF~\cite{BayTG06}. For the outlier rejection algorithms, we choose RANSAC~\cite{FischlerB81} and MAGSAC~\cite{barath2019magsac}. The errors for each category are shown in Table~\ref{table:quality} followed by the overall averaged error, where $\mathcal{I}_{3 \times 3}$ refers to a $3 \times 3$ identity matrix as a reference. In particular, feature-based methods sometimes crash, when the error is larger than $\mathcal{I}_{3 \times 3}$ error, we set the error equal to $\mathcal{I}_{3 \times 3}$ error. Regarding the motion model, from $3$) to $10$) and $12$) are single homography, $11$) is mesh-based, and $13$) is a homography array. In Table~\ref{table:quality}, we mark the best performance in red and the second-best in blue. 

As shown, except for comparing to feature-based methods in RE scenes, our method outperforms the others for all categories. It is reasonable because, in regular(RE) scenes, a set of high-quality features is detected which allows to output a good solution. In contrast, gyroscopes can only compensate for rotational motions, which decreases scores to some extent. For the rest scenes, our method beats the others with an average error being lower than the $2$nd best by $56.07\%$. Especially for low-light(LL) scenes, our method computes an error which is at least lower than the $2$nd best by $43.9\%$. 


\subsection{Ablation Studies}
\subsubsection{Fully Connected Neural Network}

\begin{figure}[h]
\begin{center}
  \includegraphics[width=1\linewidth]{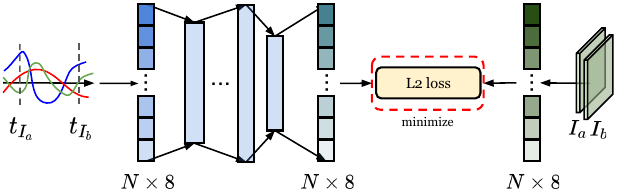}
\end{center}
  \caption{Regression of the homography array using the fully connected network. For each pair of frames $I_a$ and $I_b$, a homography array is computed by using gyro readings, which is fed to the network. On the other side, a Fundamental Mixtures model is produced as targets to guide the training process.}
\label{fig:h4pt}
\end{figure}

Our network is fully convolutional, where we convert gyroscope data into homography arrays, and then into flow fields as image input to the network. However, there is another option where we can directly input homography arrays as input. Similarly, on the other side, the Fundamental Mixtures are converted into rotation-only homography arrays and then used as guidance. Fig.~\ref{fig:h4pt} shows the pipeline, where we test two homography representations, including $3 \times 3$ homography matrix elements and $H_{4pt}$ representation from~\cite{detone2016deep} that represents homography by 4 motion vectors. The network is fully connected and L2 loss is adopted for the regression. We adopt the same training data as described above. The result is that neither representations converge, where the $H_{4pt}$ is slightly better than directly regressing matrix elements. 

Perhaps, there exist other representations or network structures that may work well or even better than our current proposal. Here, as the first try, we have proposed a working pipeline and want to leave the improvements as future works.    

\subsubsection{Global Fundamental vs. Mixtures}


To verify the effectiveness of our Fundamental Mixtures model, we compare it with a global fundamental matrix. Here, we choose the evaluation dataset of the regular scenes to alleviate the feature problem. We estimate global fundamental matrix and Fundamental Mixtures, then convert to the rotation-only homographies, respectively. Finally, we align the images with rotation-only homographies accordingly. An array of homographies from Fundamental Mixtures 
produces an error of \textbf{0.451}, which is better than an error of \textbf{0.580} produced by a single homography from a global fundamental matrix. It indicates that the Fundamental Mixtures model is functional in the case of RS cameras.  

Moreover, we generate GT with the two methods and train our network, respectively. As shown in Table~\ref{table:GT_F_Fmix}, the network trained on Fundamental Mixtures-based GT outperforms the global fundamental matrix, which demonstrates the effectiveness of our Fundamental Mixtures.
\begin{table}[h]
    \small
    \centering
    \resizebox{0.48 \textwidth}{!}
    {\begin{tabular}{lccccc}
        \toprule
        Ground Truth         & RE    & LT    & LL    & MF    & Avg   \\ \midrule
        Global Fundamental   & 0.930  &1.189 & 0.769 & 0.917 & 0.951 \\
        Fundamental Mixtures & 0.609 & 1.010  & 0.637 & 0.736 & 0.749 \\ \bottomrule
    \end{tabular}}
    \caption{The performance of networks trained on two different GT.}
    \label{table:GT_F_Fmix}
\end{table}

\subsubsection{Backbone}
\begin{table}[h]
	\centering
    \small
	\begin{tabular}{lccccc}
		\toprule  
		 &RE& LT& LL& MF& Avg \\ 
		\midrule  
		R2UNet\cite{alom2018recurrent} &0.713 &1.006 &0.652 &0.791 &0.791\\
		AttUNet\cite{oktay2018attention} &0.896 &1.058 &0.752 &0.993 &0.925\\
		R2AttUNet\cite{alom2018recurrent} &0.651 &1.014 &0.668 &0.722 &0.764\\
		\midrule  
		Ours &0.609 & 1.01 &0.637 &0.736 &0.749\\
		\bottomrule  
	\end{tabular}
	\caption{The performance of networks with different backbones.}
	\label{tab:backbone}
\end{table}
We choose the UNet~\cite{ronneberger2015u} as our network backbone, we also test several other variants~\cite{alom2018recurrent,oktay2018attention}. Except for AttUNet~\cite{oktay2018attention}, performances are similar, as shown in Table~\ref{tab:backbone}.

\section{Conclusion}
We have presented a DeepOIS pipeline for the compensation of OIS motions for gyroscope image registration. We have captured the training data as video frames as well as their gyro readings by an OIS camera and then calculated the ground-truth motions with our proposed Fundamental Mixtures model under the setting of rolling shutter cameras. For the evaluation, we have manually marked point correspondences on our captured dataset for quantitative metrics. The results show that our compensation network works well when compared with non-OIS cameras and outperforms other image-based methods. In summary, a new problem is proposed and we show that it is solvable by learning the OIS motions, such that gyroscope can be used for image registration on OIS cameras. We hope our work can inspire more researches in this direction.     


\begin{thebibliography}{10}\itemsep=-1pt

\bibitem{alom2018recurrent}
Md~Zahangir Alom, Mahmudul Hasan, Chris Yakopcic, Tarek~M Taha, and Vijayan~K
  Asari.
\newblock Recurrent residual convolutional neural network based on u-net
  (r2u-net) for medical image segmentation.
\newblock {\em arXiv preprint arXiv:1802.06955}, 2018.

\bibitem{barath2019magsac}
Daniel Barath, Jiri Matas, and Jana Noskova.
\newblock Magsac: marginalizing sample consensus.
\newblock In {\em {Proc. CVPR}}, pages 10197--10205, 2019.

\bibitem{BayTG06}
Herbert Bay, Tinne Tuytelaars, and Luc~Van Gool.
\newblock {SURF:} speeded up robust features.
\newblock In {\em {Proc. ECCV}}, volume 3951, pages 404--417, 2006.

\bibitem{BrownL03}
Matthew Brown and David~G. Lowe.
\newblock Recognising panoramas.
\newblock In {\em {Proc. ICCV}}, pages 1218--1227, 2003.

\bibitem{chiu2007optimal}
Chi-Wei Chiu, Paul C-P Chao, and Din-Yuan Wu.
\newblock Optimal design of magnetically actuated optical image stabilizer
  mechanism for cameras in mobile phones via genetic algorithm.
\newblock {\em IEEE Trans. on Magnetics}, 43(6):2582--2584, 2007.

\bibitem{dai2015euler}
Jian~S Dai.
\newblock Euler--rodrigues formula variations, quaternion conjugation and
  intrinsic connections.
\newblock {\em Mechanism and Machine Theory}, 92:144--152, 2015.

\bibitem{detone2016deep}
Daniel DeTone, Tomasz Malisiewicz, and Andrew Rabinovich.
\newblock Deep image homography estimation.
\newblock {\em arXiv preprint arXiv:1606.03798}, 2016.

\bibitem{dosovitskiy2015FlowNet}
Alexey Dosovitskiy, Philipp Fischer, Eddy Ilg, Philip H{\"{a}}usser, Caner
  Hazirbas, Vladimir Golkov, Patrick van~der Smagt, Daniel Cremers, and Thomas
  Brox.
\newblock Flownet: Learning optical flow with convolutional networks.
\newblock In {\em {Proc. ICCV}}, 2015.

\bibitem{FischlerB81}
Martin~A. Fischler and Robert~C. Bolles.
\newblock Random sample consensus: {A} paradigm for model fitting with
  applications to image analysis and automated cartography.
\newblock {\em Commun. {ACM}}, 24(6):381--395, 1981.

\bibitem{gao2011constructing}
Junhong Gao, Seon~Joo Kim, and Michael~S Brown.
\newblock Constructing image panoramas using dual-homography warping.
\newblock In {\em {Proc. CVPR}}, pages 49--56, 2011.

\bibitem{grundmann2012calibration}
Matthias Grundmann, Vivek Kwatra, Daniel Castro, and Irfan Essa.
\newblock Calibration-free rolling shutter removal.
\newblock In {\em IEEE international conference on computational photography
  (ICCP)}, pages 1--8, 2012.

\bibitem{guo2016joint}
Heng Guo, Shuaicheng Liu, Tong He, Shuyuan Zhu, Bing Zeng, and Moncef Gabbouj.
\newblock Joint video stitching and stabilization from moving cameras.
\newblock {\em IEEE Trans. on Image Processing}, 25(11):5491--5503, 2016.

\bibitem{guse2012gesture}
Dennis Guse and Benjamin M{\"u}ller.
\newblock Gesture-based user authentication for mobile devicesusing
  accelerometer and gyroscope.
\newblock In {\em Informatiktage}, pages 243--246, 2012.

\bibitem{hartley2003multiple}
Richard Hartley and Andrew Zisserman.
\newblock {\em Multiple view geometry in computer vision}.
\newblock Cambridge university press, 2003.

\bibitem{huang2018online}
Weibo Huang and Hong Liu.
\newblock Online initialization and automatic camera-imu extrinsic calibration
  for monocular visual-inertial slam.
\newblock In {\em IEEE International Conference on Robotics and Automation
  (ICRA)}, pages 5182--5189, 2018.

\bibitem{jia2013online}
Chao Jia and Brian~L Evans.
\newblock Online calibration and synchronization of cellphone camera and
  gyroscope.
\newblock In {\em IEEE Global Conference on Signal and Information Processing},
  pages 731--734, 2013.

\bibitem{karpenko2011digital}
Alexandre Karpenko, David Jacobs, Jongmin Baek, and Marc Levoy.
\newblock Digital video stabilization and rolling shutter correction using
  gyroscopes.
\newblock {\em CSTR}, 1(2011):2, 2011.

\bibitem{kingma2014adam}
Diederik~P Kingma and Jimmy Ba.
\newblock Adam: A method for stochastic optimization.
\newblock {\em arXiv preprint arXiv:1412.6980}, 2014.

\bibitem{koo2009optical}
Jun-Mo Koo, Myoung-Won Kim, and Byung-Kwon Kang.
\newblock Optical image stabilizer for camera lens assembly, Feb.~10 2009.
\newblock US Patent 7,489,340.

\bibitem{le2020deep}
Hoang Le, Feng Liu, Shu Zhang, and Aseem Agarwala.
\newblock Deep homography estimation for dynamic scenes.
\newblock In {\em {Proc. CVPR}}, pages 7652--7661, 2020.

\bibitem{lin2017direct}
Kaimo Lin, Nianjuan Jiang, Shuaicheng Liu, Loong-Fah Cheong, Minh~N Do, and
  Jiangbo Lu.
\newblock Direct photometric alignment by mesh deformation.
\newblock In {\em {Proc. CVPR}}, pages 2701--2709, 2017.

\bibitem{LiuTYSZ16}
Shuaicheng Liu, Ping Tan, Lu Yuan, Jian Sun, and Bing Zeng.
\newblock Meshflow: Minimum latency online video stabilization.
\newblock In {\em {Proc. ECCV}}, volume 9910, pages 800--815, 2016.

\bibitem{liu2017hybrid}
Shuaicheng Liu, Binhan Xu, Chuang Deng, Shuyuan Zhu, Bing Zeng, and Moncef
  Gabbouj.
\newblock A hybrid approach for near-range video stabilization.
\newblock {\em IEEE Trans. on Circuits and Systems for Video Technology},
  27(9):1922--1933, 2017.

\bibitem{liu2013bundled}
Shuaicheng Liu, Lu Yuan, Ping Tan, and Jian Sun.
\newblock Bundled camera paths for video stabilization.
\newblock {\em {ACM Trans. Graphics}}, 32(4), 2013.

\bibitem{Lowe04}
David~G. Lowe.
\newblock Distinctive image features from scale-invariant keypoints.
\newblock {\em Int. J. Comput. Vis.}, 60(2):91--110, 2004.

\bibitem{mustaniemi2019gyroscope}
Janne Mustaniemi, Juho Kannala, Simo S{\"a}rkk{\"a}, Jiri Matas, and Janne
  Heikkila.
\newblock Gyroscope-aided motion deblurring with deep networks.
\newblock In {\em 2019 IEEE Winter Conference on Applications of Computer
  Vision (WACV)}, pages 1914--1922. IEEE, 2019.

\bibitem{nasiri2012optical}
Steven~S Nasiri, Mansur Kiadeh, Yuan Zheng, Shang-Hung Lin, and SHI Sheena.
\newblock Optical image stabilization in a digital still camera or handset,
  May~1 2012.
\newblock US Patent 8,170,408.

\bibitem{oktay2018attention}
Ozan Oktay, Jo Schlemper, Loic~Le Folgoc, Matthew Lee, Mattias Heinrich,
  Kazunari Misawa, Kensaku Mori, Steven McDonagh, Nils~Y Hammerla, Bernhard
  Kainz, et~al.
\newblock Attention u-net: Learning where to look for the pancreas.
\newblock {\em arXiv preprint arXiv:1804.03999}, 2018.

\bibitem{EpicFlow_2015}
Jerome Revaud, Philippe Weinzaepfel, Zaid Harchaoui, and Cordelia Schmid.
\newblock Epicflow: Edge-preserving interpolation of correspondences for
  optical flow.
\newblock In {\em {Proc. CVPR}}, pages 1164--1172, 2015.

\bibitem{ronneberger2015u}
Olaf Ronneberger, Philipp Fischer, and Thomas Brox.
\newblock U-net: Convolutional networks for biomedical image segmentation.
\newblock In {\em International Conference on Medical image computing and
  computer-assisted intervention}, pages 234--241. Springer, 2015.

\bibitem{rublee2011orb}
Ethan Rublee, Vincent Rabaud, Kurt Konolige, and Gary Bradski.
\newblock Orb: An efficient alternative to sift or surf.
\newblock In {\em {Proc. ICCV}}, pages 2564--2571, 2011.

\bibitem{RubleeRKB11}
Ethan Rublee, Vincent Rabaud, Kurt Konolige, and Gary~R. Bradski.
\newblock {ORB:} an efficient alternative to {SIFT} or {SURF}.
\newblock In {\em {Proc. ICCV}}, pages 2564--2571, 2011.

\bibitem{sato1993control}
Koichi Sato, Shigeki Ishizuka, Akira Nikami, and Mitsuru Sato.
\newblock Control techniques for optical image stabilizing system.
\newblock {\em IEEE Trans. on Consumer Electronics}, 39(3):461--466, 1993.

\bibitem{shan2007rotational}
Qi Shan, Wei Xiong, and Jiaya Jia.
\newblock Rotational motion deblurring of a rigid object from a single image.
\newblock In {\em 2007 IEEE 11th International Conference on Computer Vision},
  pages 1--8. IEEE, 2007.

\bibitem{shi1994good}
Jianbo Shi et~al.
\newblock Good features to track.
\newblock In {\em 1994 Proceedings of IEEE conference on computer vision and
  pattern recognition}, pages 593--600. IEEE, 1994.

\bibitem{siddha2012hardware}
Vividh Siddha, Kunihiro Ishiguro, and Guillermo~A Hernandez.
\newblock Hardware abstraction layer, Aug.~28 2012.
\newblock US Patent 8,254,285.

\bibitem{sun2018pwc}
Deqing Sun, Xiaodong Yang, Ming-Yu Liu, and Jan Kautz.
\newblock Pwc-net: Cnns for optical flow using pyramid, warping, and cost
  volume.
\newblock In {\em {Proc. CVPR}}, pages 8934--8943, 2018.

\bibitem{TianYFWHB19}
Yurun Tian, Xin Yu, Bin Fan, Fuchao Wu, Huub Heijnen, and Vassileios Balntas.
\newblock Sosnet: Second order similarity regularization for local descriptor
  learning.
\newblock In {\em {Proc. CVPR}}, pages 11016--11025, 2019.

\bibitem{tian2019sosnet}
Yurun Tian, Xin Yu, Bin Fan, Fuchao Wu, Huub Heijnen, and Vassileios Balntas.
\newblock Sosnet: Second order similarity regularization for local descriptor
  learning.
\newblock In {\em {Proc. CVPR}}, pages 11016--11025, 2019.

\bibitem{trajkovic1998fast}
Miroslav Trajkovi{\'c} and Mark Hedley.
\newblock Fast corner detection.
\newblock {\em Image and vision computing}, 16(2):75--87, 1998.

\bibitem{WronskiGEKKLLM19}
Bartlomiej Wronski, Ignacio Garcia{-}Dorado, Manfred Ernst, Damien Kelly,
  Michael Krainin, Chia{-}Kai Liang, Marc Levoy, and Peyman Milanfar.
\newblock Handheld multi-frame super-resolution.
\newblock {\em {ACM Trans. Graphics}}, 38(4):28:1--28:18, 2019.

\bibitem{yeom2009optical}
DH Yeom.
\newblock Optical image stabilizer for digital photographing apparatus.
\newblock {\em IEEE Trans. on Consumer Electronics}, 55(3):1028--1031, 2009.

\bibitem{YiTLF16}
Kwang~Moo Yi, Eduard Trulls, Vincent Lepetit, and Pascal Fua.
\newblock {LIFT:} learned invariant feature transform.
\newblock In {\em {Proc. ECCV}}, volume 9910, pages 467--483, 2016.

\bibitem{zaki2020study}
Tirra Hanin~Mohd Zaki, Musab Sahrim, Juliza Jamaludin, Sharma~Rao Balakrishnan,
  Lily~Hanefarezan Asbulah, and Filzah~Syairah Hussin.
\newblock The study of drunken abnormal human gait recognition using
  accelerometer and gyroscope sensors in mobile application.
\newblock In {\em 2020 16th IEEE International Colloquium on Signal Processing
  \& Its Applications (CSPA)}, pages 151--156, 2020.

\bibitem{zaragoza2013projective}
Julio Zaragoza, Tat-Jun Chin, Michael~S Brown, and David Suter.
\newblock As-projective-as-possible image stitching with moving dlt.
\newblock In {\em {Proc. CVPR}}, pages 2339--2346, 2013.

\bibitem{zhang2020content}
Jirong Zhang, Chuan Wang, Shuaicheng Liu, Lanpeng Jia, Jue Wang, Ji Zhou, and
  Jian Sun.
\newblock Content-aware unsupervised deep homography estimation.
\newblock In {\em {Proc. ECCV}}, 2020.

\end{thebibliography}

\end{document}